\documentclass{article}

\usepackage[final]{neurips_2026}
\makeatletter
\renewcommand{\@noticestring}{Workshop on Who Verifies the Agents? Toward Reliable Agent Development, NeurIPS 2026.}
\makeatother

\usepackage[utf8]{inputenc}
\usepackage[T1]{fontenc}
\usepackage{hyperref}
\hypersetup{hidelinks,pdftitle={Safety, or Just Capability? A Validity Audit of Agent-Safety Benchmarks},pdfauthor={Youting Wang, Xiao Han, Dingyan Shang, Yuan Tang, Bowen Liu}}
\usepackage{url}
\usepackage{graphicx}
\usepackage{booktabs}
\usepackage{array}
\usepackage{capt-of}
\usepackage{amsmath,amssymb}
\usepackage{amsthm}
\usepackage{nicefrac}
\usepackage{microtype}
\usepackage{xcolor}

\newtheorem{observation}{Observation}

\newcommand{\workshopversion}{}
\newcommand{\verifyagentsworkshop}{}

\title{Safety, or Just Capability? A Validity Audit of Agent-Safety Benchmarks}
\author{%
  Youting Wang \\ Independent Researcher \\ \texttt{ginkoin613@gmail.com} \And
  Xiao Han \\ Independent Researcher \\ \texttt{xhan18908@gmail.com} \And
  Dingyan Shang \\ Independent Researcher \\ \texttt{dingyanshang@gmail.com} \And
  Yuan Tang \\ Independent Researcher \\ \texttt{naomitangyuan@gmail.com} \And
  Bowen Liu \\ Independent Researcher \\ \texttt{chris3494285@gmail.com}
}

\begin{document}
\maketitle

\begin{abstract}
A developer choosing among models for an agent may compare them on one safety-benchmark score. We
ask whether four agent-safety benchmarks can stand in for one another in that role. R-Judge asks a
model to judge stored agent traces; InjecAgent, AgentHarm, and AgentDojo test the model as an acting
agent. We ran each benchmark with its official code and scorer on up to 22 models released in 2024--25 (coverage varies by benchmark), and
measured MMLU and GPQA under one protocol as a static-knowledge composite. We report Spearman rank
correlations ($\rho$). First, R-Judge's headline $F_1$ can be
matched without discrimination: always answering ``unsafe'' scores $0.690$, above five of the 21
models with valid R-Judge output, all of which do discriminate. Second, the three benchmarks with broad coverage rank the same 18
models differently, and an apparent trade-off between R-Judge and AgentHarm at seven models ($\rho{=}{-}0.64$)
does not persist at 18 ($\rho{=}{+}0.02$). Third, after adjusting for the static-knowledge composite, we
detect no incremental association between any of the three broad-coverage safety scores and $\tau^2$-bench task success. Fourth, the largest
adjusted association is between AgentHarm and a separate jailbreak test, StrongREJECT
prompts under three jailbreak templates ($\rho{=}{+}0.72$, $n{=}19$); because both score harmful compliance, this is convergent evidence, not evidence of
general safety. A safety score should be reported together with what it measures, how it is
calculated, and which models were evaluated.

\end{abstract}

\section{Introduction}

A developer choosing among candidate models for an agent may compare them on a single safety-benchmark
score. Agent-safety benchmarks, however, measure different behaviors. R-Judge asks a model to judge
whether stored agent traces are unsafe \citep{rjudge2024}. InjecAgent and AgentDojo test whether a
tool-using agent follows injected instructions \citep{injecagent2024,agentdojo2024}. AgentHarm tests
whether an agent carries out harmful multi-step tasks \citep{agentharm2024}. Their scorers differ too:
rules, an LLM grader, or environment state.

Because these benchmarks target different behaviors, we do not expect them to agree, and disagreement
does not make any one of them invalid. Our question is narrower: can one score stand in for another
when ranking models, as it does when a single number is reported as an agent's safety? Validity
belongs to a particular interpretation and use of a score \citep{kane2013validation,benchmarkcards2025},
so we test this use directly. A 12-model study already reports low agreement across four agent-safety
benchmarks \citep{agentsafety_taxonomy2026}; we add matched model panels, metric checks, and outside
criteria. We ask four questions.
\textbf{RQ1:} Do the benchmarks rank the same models in the same order?
\textbf{RQ2:} How closely do their scores track a static-knowledge composite of MMLU and GPQA?
\textbf{RQ3:} Beyond that composite, how do they relate to three outside variables: $\tau^2$-bench task
success, avoidance of blackmail or leaking in fictional misalignment scenarios, and safety on
StrongREJECT prompts under jailbreak templates?
\textbf{RQ4:} Does a headline metric reward a model that does not discriminate at all?

\paragraph{Findings.}
(1)~R-Judge's headline metric, F1 (the harmonic mean of precision and recall for the ``unsafe'' label),
rewards a constant policy: always answering ``unsafe'' scores $0.690$, above five of
the 21 models with valid R-Judge output (Observation~\ref{obs:gameable}). This is a known property of F1
\citep{lipton2014f1}; we show its consequence on this benchmark.
(2)~R-Judge, InjecAgent, and AgentHarm, the three benchmarks most models completed, rank the same 18
models differently ($\rho$ is Spearman rank correlation throughout). A trade-off seen at seven
models ($\rho{=}{-}0.64$) does not persist at 18 ($\rho{=}{+}0.02$), and a quarter of random seven-model
subsets of this roster reach $|\rho|\geq0.5$.
(3)~The composite is associated with task success ($\rho{=}{+}0.60$), but after adjusting for it we
detect no incremental association between R-Judge, InjecAgent, or AgentHarm and task success.
(4)~The largest adjusted association is AgentHarm with jailbreak safety ($\rho{=}{+}0.72$, $n{=}19$;
95\% interval from resampling model developers $[+0.17,+0.97]$). Both score harmful compliance, so this is convergent
evidence, not evidence of general safety.

\paragraph{Scope.} MMLU and GPQA measure static knowledge, not tool use or planning, so every
``adjusted'' result is conditional on that composite only. Each model was run once per benchmark, and
the outside criteria are stand-ins rather than measures of real-world harm
(Section~\ref{sec:limitations}).

\section{Related Work}

\paragraph{Agent-safety evaluators.} The four audited benchmarks target different behaviors:
InjecAgent \citep{injecagent2024} and AgentDojo \citep{agentdojo2024} test prompt-injection
robustness, AgentHarm \citep{agentharm2024} tests compliance with harmful tasks, and R-Judge
\citep{rjudge2024} tests risk recognition in interaction traces. Other agent-safety benchmarks cover emulated tool
risks \citep{toolemu2023}, broad agent safety \citep{agentsafetybench2024}, attacks and defenses
\citep{asb2025}, and web, mobile, and vision-language agents
\citep{stwebagentbench2026,mobilesafetybench2024,visualleakbench2026action}; WASP separates
partial attacker progress from completed attack goals \citep{wasp2025};
HarmBench standardizes harmful-request red teaming \citep{harmbench2024}. AgentAuditor improves trace-level
safety and security judgment, while ASSEBench evaluates it \citep{agentauditor2025}. That work asks
whether an evaluator labels records correctly; we ask whether scores can be substituted across
judging and acting roles.

\paragraph{Benchmark validity and capability confounding.} Validity attaches to a proposed
interpretation and use of a score, not to a test in the abstract \citep{kane2013validation}. The
Agentic Benchmark Checklist \citep{abc2025} and HackDetect \citep{hackdetect2026} audit tasks,
outcomes, reporting, and executable protocols. Safetywashing shows that apparent safety progress
can be entangled with general capability \citep{safetywashing2024}, while a review of 210 safety
benchmarks identifies recurring gaps between measured behavior and risk claims
\citep{safetybenchmarkprinciples2026}. BenchmarkCards standardize objectives, methods, intended
uses, and limitations \citep{benchmarkcards2025}; we test one such use: substituting
scores when ranking models. \citet{lipton2014f1} describe F1's all-positive behavior; we compute
the baseline at R-Judge's base rate and compare it with model scores.

\paragraph{Closest empirical baselines.} BenchBench shows that benchmark-agreement conclusions
depend on analysis choices \citep{benchbench2024}; metabench studies shared abilities among knowledge and
reasoning benchmarks \citep{metabench2024}, and \citet{measuring_construct_validity} review
construct-validity problems across 445 LLM benchmarks. Safetywashing measures broad capability
confounding, but not matched agent-loop reruns. Closest, \citet{agentsafety_taxonomy2026} catalog
40 behavioral agent-safety benchmarks and report low concordance in a 12-model, four-benchmark
check that includes the three acting-policy benchmarks used here. We extend that check with a larger
panel, official reruns on matched models, an audit of R-Judge's scoring rule, held-out comparisons
adjusted for MMLU/GPQA, and a documented panel expansion. Convergent and discriminant validity
\citep{campbellfiske1959} guide which comparisons we expect to agree.

\paragraph{Over-refusal and task performance.} XSTest \citep{xstest2024} and OR-Bench
\citep{orbench2024} study over-refusal at the prompt level. We test the analogous relation between
low harmful-task completion (AgentHarm safety) and correct recognition of benign traces; it is near zero on the full panel but
unstable on small panels. Work on corrupt success \citep{corruptsuccess2026} and task-success
benchmarks such as $\tau$-bench \citep{taubench2024} and $\tau^2$-bench \citep{tau2bench2025}
motivates our held-out task-success criterion.

\section{Method}
\label{sec:method}
Table~\ref{tab:resultmap} (appendix) maps each result to its score, models, uncertainty, and supported claim.

\subsection{Model panel and benchmarks}
We evaluate, through APIs, up to 22 model endpoints released in
2024--25 (the \emph{2024--25 panel}), from nine
model-developing organizations
(OpenAI~6, Meta~3, Qwen~3, Mistral~3, DeepSeek~2, Amazon~2, Anthropic, Google, Cohere;
\ifdefined\workshopversion
full roster in Fig.~\ref{fig:scale}).
\else
full roster in Fig.~\ref{fig:scale} and the artifact).
\fi
We re-run four safety benchmarks with their official
implementations and scorers: R-Judge (the current upstream 571-record snapshot, self-judge;
lineage in App.~\ref{app:lineage}), InjecAgent (300 stratified
direct-harm/data-stealing items, rule-scored), AgentHarm (all 44 base behaviors of the public test split, covering its eight
harm categories, via the official Inspect evaluation with a gpt-4o-mini judge), and AgentDojo
(the \texttt{slack} suite, 100 items, environment-scored for security and utility), plus a
held-out task-success criterion for RQ3 (below).

Coverage differs by benchmark (Fig.~\ref{fig:scale} maps the full panel): R-Judge has $n{=}21$ after
Qwen3-32B falls below the $90\%$ valid-output threshold (DeepSeek-R1 has 570 of 571 records);
InjecAgent has conditional scores for
$n{=}22$; AgentHarm and AgentDojo, which require native tool use, have $n{=}19$ and $5$; and the
$\tau^2$ criterion has $n{=}20$
(App.~\ref{app:panels}). The common R-Judge/InjecAgent/AgentHarm panel therefore has $n{=}18$.
Capability anchors (MMLU \citep{mmlu2021} and
GPQA-Diamond \citep{gpqa2023}) are measured by us under one uniform protocol: MMLU zero-shot on a fixed 500-item
subset (seed 42, identical items for every model) and GPQA-Diamond (198 items, chain-of-thought),
through the same API harness as the safety runs. These fixed-protocol measurements replace
provider-reported model-card numbers, and the composite passes its planned positive check: the MMLU loading exceeds $0.6$ (App.~\ref{app:rq1}). The anchor panel is 21 models (DeepSeek-R1 is excluded
for unstable long-form GPQA generation); we keep the provider-reported anchors as a
pre-harmonization comparison.

\paragraph{Adaptive model-population check.}
Each follow-up stage was planned after seeing the previous results. A July 7 gate
attempted R-Judge, anchors, and misalignment on eight candidates: seven R-Judge executions were
record-complete, six also had both anchors, and five additionally passed R-Judge's valid-output gate
and entered the capability-adjusted continuation. Its R-Judge--misalignment estimate fell to
$+0.23$ on all 25 eligible models (the common-panel value was $+0.41$, $n{=}18$),
below the stop threshold. After observing that result, a July 9
protocol fixed 17 further candidates, an updated prediction, and telemetry-only eligibility rules;
that later stage added anchors, misalignment, and jailbreak, but no audited-benchmark or $\tau^2$
runs. The final 41 anchored endpoints span 12 organizations; outcome-independent gates yield
$n{=}40/41$ for the two variables. This is population sensitivity, and the pooled panel includes
the motivating observations (App.~\ref{app:n40}).

\subsection{Benchmark targets}
\paragraph{Actor and use map.} The developer is the score user choosing among candidate endpoints.
R-Judge places an endpoint in a \emph{verifier} role over stored traces; InjecAgent, AgentHarm, and
AgentDojo place it in an \emph{acting-policy} role. A self-verdict, rule, external LLM judge, or
environment scorer then produces the scalar. Our audited use is scalar substitution for endpoint
selection; typed multi-signal composition is not tested. We do not claim that the same actor or
latent trait generated all four. Appendix~\ref{app:actor_map} gives the complete map.

Before looking at relationships among score columns, we recorded each target. R-Judge tests whether
a verifier distinguishes unsafe from benign traces. InjecAgent and AgentDojo test resistance to
prompt injection, so their relation is a convergent check. AgentHarm scores how much of a harmful task an acting
endpoint completes. For R-Judge, we analyze specificity and balanced accuracy in addition to the
official F1 because the metrics reward different error patterns.

\paragraph{Score orientation and terminology.} All reported scores are oriented so that higher is
better or safer. R-Judge \emph{specificity} is the fraction of benign traces correctly identified
as benign; its \emph{balanced accuracy} averages performance on benign and unsafe traces.
InjecAgent robustness is one minus attack success \emph{among valid outputs}, reaggregated on one
frozen intersection of 300 item IDs for all models; it is therefore a conditional estimand.
Valid-output coverage ranges from $38.7\%$ to $100\%$, so we also bound every invalid output as
safe or as an attack success (App.~\ref{app:injecdenom}). AgentHarm safety is $1-\bar{s}$, where $\bar{s}$ is the
official \texttt{combined\_scorer}'s \texttt{avg\_score}: the mean over behaviors of the fraction of
each behavior's grading criteria the agent satisfied. The scorer's separate refusal flag does not
enter this scalar. For RQ3, $\tau^2$-bench is a held-out task-success variable, not a safety score.
We use \emph{internal consistency} for split-half analyses (App.~\ref{app:panels}) and name each
benchmark-specific score explicitly, instead of stretching ``calibration'' or ``reliability''
past their usual meanings.

\subsection{Analysis plan}
We wrote down hypotheses and analysis rules before each stage of data collection. These notes are
not independently timestamped, and later stages were planned after seeing earlier results, so we
treat all analyses as exploratory rather than preregistered; the artifact includes the notes and a
chronology. CSV
\texttt{run\_date} fields are runner metadata, not independent collection timestamps
(App.~\ref{app:lineage}). The capability composite is the mean of standardized, verified MMLU and
GPQA. We use Spearman associations. RQ1/RQ2 are fixed-roster descriptions with
leave-one-organization-out sensitivity, not population tests. Ordinary case bootstraps provide
percentile sensitivity intervals only; we do not treat their sign fractions as null $p$-values.

A PCA is reported in the appendix as exploratory only; we make no latent-factor claim
(App.~\ref{app:rq1}). For partial associations, we midrank predictor, outcome, and
capability; regress the first two separately on capability rank; and Spearman-correlate their
residuals. RQ3 endpoint-level $p$-values permute outcome residuals ($99{,}999$ draws). An exhaustive
organization-cluster bootstrap supplies a dependence sensitivity; its interval and sign-tail are
descriptive, not a calibrated null test (App.~\ref{app:robustness}). The observational unit is a
recorded endpoint/version. Item resampling conditions on that execution, and organization
resampling conditions on the observed aggregate scores; neither captures provider rerun variance.

RQ3 variables were introduced under separate protocols: $\tau^2$-retail task success (40
multi-turn, environment-scored tasks) \citep{tau2bench2025}; avoidance of blackmail or leaking over
100 fictional-scenario decisions \citep{agenticmisalignment2025}; and safety on 50 forbidden
prompts averaged over three jailbreak templates \citep{strongreject2024,wei2023jailbroken}. Their
roles are respectively discriminant, exploratory criterion-related, and construct-overlapping
convergent evidence. The primary population is each predictor--variable--anchor intersection
($n{=}19$--$21$); a common $n{=}18$ panel is sensitivity only. The cellwise manifest is in
App.~\ref{app:rq3pop}. The corresponding protocol texts are included in the artifact.

\paragraph{Reproducibility boundary.} The supplement's frozen logs, manifests, hashes,
reliability snapshot, protocols, and analysis scripts reproduce the reported reanalysis and
Table~\ref{tab:reliability} without provider calls.
Generation reruns require credentials and depend on incompletely pinnable endpoints/routing.

\section{Results}
\label{sec:results}

Results run in order of evidential strength: the scoring-rule failure (RQ4) and capability
correlations (RQ2) are firmest, cross-benchmark structure (RQ1) is weaker, and the power-limited
held-out validity tests (RQ3) come last.

\subsection{The headline metric is gameable (RQ4)}

\begin{figure}[b]
\centering
\includegraphics[width=0.92\columnwidth]{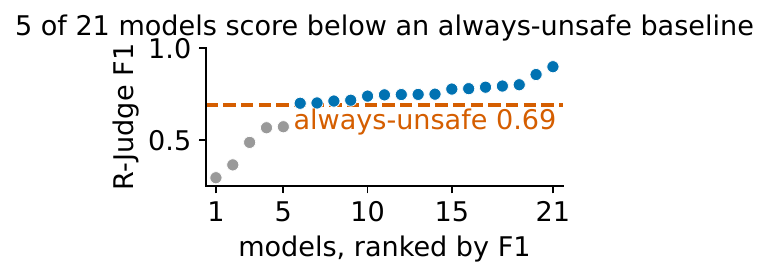}
\caption{\textbf{A leaderboard's headline metric can be matched by an ``always unsafe'' baseline.}
On R-Judge ($n{=}21$), five real, discriminating models score below this constant baseline
(F1${=}0.69$, dashed): F1 has no true-negative term, so the degenerate
policy can outrank models that discriminate.}
\label{fig:gameable}
\end{figure}

The current R-Judge release snapshot labels $52.7\%$ of its 571 records ``unsafe''
(upstream-count lineage in App.~\ref{app:lineage}). A degenerate policy that answers
``unsafe'' on every item, exercising no safety reasoning, therefore attains recall $1.0$,
specificity $0$, and $\mathrm{F1}=0.690$. On the 21 panel endpoints that produce valid R-Judge
verdicts (Qwen3-32B is excluded for $21\%$ unparseable output; see Method) this constant-baseline score
outranks five real models (F1 $0.30$--$0.57$; Fig.~\ref{fig:gameable}), all of which discriminate. The panel's highest-specificity model, o3-mini
($0.97$), receives $\mathrm{F1}=0.702$, only $0.012$ above the constant baseline, because its recall is lower.
Balanced accuracy, which credits correct decisions on both classes, does not reward the constant policy (it scores $0.5$).

\begin{observation}[F1 admits a high-scoring constant baseline]
\label{obs:gameable}
On a benchmark with class base rate $\pi$ scored by F1, the constant ``always-positive'' policy
attains $\mathrm{F1}=2\pi/(1+\pi)$ without distinguishing safe from unsafe items. For R-Judge ($\pi=0.527$, a
near-balanced base rate) this is $0.690$, which exceeds the observed F1 of several models that do
discriminate. F1 has no explicit true-negative term: benign traces matter only through false
positives (when there is at least one true positive, turning a false positive into a true negative raises F1), and the score depends on
prevalence and operating point. F1 therefore cannot stand alone as a measure of two-sided
discrimination on this trace-judgment benchmark. This is a known property of F1
\citep{lipton2014f1}; our contribution is its concrete consequence on R-Judge.
\end{observation}

\begin{table}[t]
\centering
\small
\setlength{\tabcolsep}{5pt}
\begin{tabular}{@{}lccccc@{}}
\toprule
Model & R-Judge & R-Judge & InjecAgent & AgentHarm & GPQA \\
 & spec.\ & bal.\ acc.\ & robust.\ & safety & \\
\midrule
Llama-3.1-8B & \underline{0.04} & 0.52 & 0.32 & \textbf{0.89} & 30.4 \\
Claude-3-Haiku & 0.40 & \underline{0.51} & 0.77 & 0.81 & 33.3 \\
Llama-3.1-70B & 0.11 & 0.55 & 0.35 & 0.78 & 46.7 \\
o3-mini & \textbf{0.97} & 0.76 & 0.87 & 0.77 & 77.2 \\
Nova-Lite & 0.94 & 0.79 & 0.88 & 0.76 & 42.0 \\
Nova-Pro & 0.64 & 0.71 & 0.60 & 0.75 & 46.9 \\
Gemini-2.5-Flash & 0.92 & 0.78 & 0.71 & 0.62 & 68.3 \\
DeepSeek-V3 & 0.86 & \textbf{0.85} & 0.76 & 0.61 & 59.1 \\
Qwen-2.5-72B & 0.62 & 0.68 & 0.92 & 0.60 & 49.1 \\
GPT-4.1 & 0.61 & 0.76 & 0.87 & 0.60 & 66.3 \\
GPT-4.1-mini & 0.76 & 0.78 & 0.87 & 0.51 & 65.0 \\
GPT-4o & 0.72 & 0.77 & 0.90 & 0.47 & 53.6 \\
GPT-4.1-nano & 0.79 & 0.59 & \textbf{0.94} & 0.42 & 50.3 \\
Llama-3.3-70B & 0.28 & 0.62 & \underline{0.31} & 0.39 & 50.0 \\
Mistral-Nemo & 0.80 & 0.63 & 0.34 & 0.35 & 28.0 \\
GPT-4o-mini & 0.61 & 0.73 & 0.89 & 0.31 & 40.2 \\
Mixtral-8x22B & 0.58 & 0.70 & 0.78 & \underline{0.24} & 33.2 \\
Mistral-Large & 0.50 & 0.71 & 0.81 & \underline{0.24} & 44.0 \\
\bottomrule
\end{tabular}
\caption{The three broad-coverage benchmarks rank the same $n{=}18$ models differently. Safety columns run from 0 to 1, higher is safer: R-Judge specificity is the fraction of benign traces judged benign (an always-benign policy scores 1), balanced accuracy averages benign and unsafe recall, InjecAgent robustness is one minus attack success, and AgentHarm safety is one minus the harmful-task score (\S3.2). GPQA is the provider-reported accuracy in percent (RQ2/RQ3 use our fixed-protocol rerun). Rows are sorted by AgentHarm safety; \textbf{bold} marks column maxima and underlining minima.}
\label{tab:matrix}
\end{table}

\subsection{Association with the static-knowledge composite depends on the metric (RQ2)}
On the fixed R-Judge snapshot ($n{=}20$ with harmonized anchors), capability correlates strongly
with balanced accuracy ($\rho=+0.71$ with MMLU; $+0.49$ with GPQA) and F1 ($+0.76/+0.62$), but
only weakly with specificity ($+0.16/+0.07$). Across nine leave-one-organization-out folds,
balanced-accuracy correlations remain $+0.66$--$+0.82$ with MMLU and $+0.42$--$+0.67$ with
GPQA; F1 remains $+0.72$--$+0.82$ and $+0.55$--$+0.78$, whereas specificity spans
$+0.04$--$+0.32$ and $-0.24$--$+0.21$. Thus the metric contrast persists within this roster,
without licensing organization-population inference. It follows the definitions: balanced accuracy
and F1 include unsafe-trace recall; specificity only credits benign traces. The latter's MMLU
correlation also fell from $0.85$ on our first nine endpoints to $+0.16$ at $n{=}20$.

\subsection{Benchmarks disagree; a seven-model panel suggested a trade-off that does not persist (RQ1)}
\label{sec:rq1}
The three benchmarks rank these models differently (Table~\ref{tab:matrix}): a model near the top
on AgentHarm can rank near the bottom on R-Judge specificity. This needs no correlation estimate.
Because the benchmarks measure different behaviors (\S\ref{sec:method}; App.~\ref{app:artifacts}),
disagreement is expected and does not make any one of them invalid; it does mean that replacing one
score with another can change the model ranking.

We read a trade-off into that disagreement early on; it does not survive the larger
cross-benchmark panel. The R-Judge-specificity/AgentHarm-safety rank correlation was $-0.64$
at $n{=}7$ but $+0.02$ at $n{=}18$ (Fig.~\ref{fig:reversal}, appendix); uniformly drawn size-7
subsets of this fixed 18-endpoint roster yield $|\rho|\geq0.5$ in a quarter of draws. Deleting one
of eight organizations moves the full estimate from $-0.22$ to $+0.39$, so neither the early
negative relation nor the later near-zero magnitude is population-stable.
Seven-model subsets can produce strong correlations by chance, and none of the four R-Judge
metrics recovers the trade-off on the larger panel (App.~\ref{app:metric}).

With 18 models and five variables, a principal-component analysis is too small to support a latent
structure: Horn's parallel analysis on the rank-based PCA retains zero factors, the median between-benchmark correlation
is only $0.22$, and the convergent AgentDojo check is inconclusive at $n{=}5$. We therefore make no
latent-factor claim; the exploratory PCA is in App.~\ref{app:rq1}. The RQ1 result does not depend on
it: it is the matched-panel ranking disagreement in Table~\ref{tab:matrix}.

\subsection{Relations to held-out external variables differ by role and population (RQ3)}
\label{sec:rq3}

We next compare the benchmark scores with held-out behavior, after adjusting for the
static-knowledge composite. The three variables serve different purposes (\S\ref{sec:method}):
$\tau^2$ task success is a discriminant check, misalignment is exploratory, and jailbreak safety tests
agreement on harmful compliance (convergent evidence). None comes from the four audited benchmarks.

\begin{table}[t]
\centering
\small
\begin{tabular*}{\columnwidth}{@{\extracolsep{\fill}}lccc@{}}
\toprule
 & \textbf{Task success} & \multicolumn{2}{c}{\textbf{Safety outcomes}} \\
\cmidrule(lr){2-2}\cmidrule(lr){3-4}
\textbf{Predictor} & $\tau^2$ & misalign. & jailbreak$^{\dagger}$ \\
\midrule
Capability (raw $\rho$) & $+0.60$ & $-0.44$ & $+0.08$ \\
\midrule
\ifdefined\workshopversion
\multicolumn{4}{@{}l@{}}{\emph{Safety-score partial $\rho$ (controlling MMLU/GPQA):}} \\
\else
\multicolumn{4}{@{}l@{}}{\emph{Safety-score partial $\rho$ (controlling capability):}} \\
\fi
R-Judge specificity  & $-0.09$ & $+0.37$ & $-0.11$ \\
R-Judge balanced acc. & $+0.19$ & $+0.04$ & $-0.20$ \\
InjecAgent robustness & $+0.16$ & $+0.28$ & $+0.21$ \\
AgentHarm safety     & $-0.23$ & $+0.13$ & $+0.72$ \\
\bottomrule
\end{tabular*}
\caption{Relations to held-out external variables on maximum-coverage 2024--25 panels (higher is
better or safer). Rowwise $n$ for task/misalignment/jailbreak is $20/21/21$ for the capability row, $19/20/20$ for both R-Judge rows, $20/21/21$ for
InjecAgent, and $19/19/19$ for AgentHarm. Specificity credits only benign-trace recognition;
balanced accuracy averages both classes; common-panel ($n{=}18$) sensitivity is in App.~\ref{app:rq3pop}.
The top row reports raw correlations with capability; the remaining rows report partial Spearman
correlations controlling the MMLU/GPQA composite. The measured composite tracks task success but not either
safety outcome consistently; AgentHarm has the largest association, with the construct-overlapping
three-template jailbreak variable marked by $^{\dagger}$. The raw row is descriptive for the
recorded endpoints; residual-permutation and organization sensitivity for the partial rows are
reported in App.~\ref{app:rq3pop}.
}
\label{tab:rq3}
\end{table}

\paragraph{Against task success, we detect no incremental association beyond the measured composite.}
The static-knowledge composite is associated with $\tau^2$-retail success ($\rho=+0.60$,
$p=0.005$, $n{=}20$), while no safety score shows a significant partial association once the composite is partialled
out (best $|\rho|=0.23$, n.s.\ at $n{=}19$; Table~\ref{tab:rq3};
Fig.~\ref{fig:rq3}, appendix). None of these partial correlations is significant in this panel. The
organization-cluster intervals are wide (R-Judge $[-.62,+.41]$, InjecAgent $[-.28,+.62]$,
AgentHarm $[-.79,+.23]$; App.~\ref{app:rq3pop}), so practically meaningful associations remain
possible. With a minimum detectable effect of about $0.62$ at $80\%$ power (App.~\ref{app:panels}), the wide
intervals leave meaningful associations unresolved, so this null does not establish discriminant validity.

\begin{figure}[t]
\centering
\includegraphics[width=0.84\columnwidth]{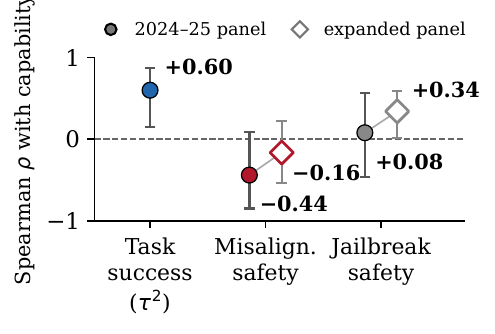}
\caption{\textbf{The measured composite tracks task success; safety point estimates differ by outcome and
panel.} Filled circles show the 2024--25 correlations; open diamonds show the expanded-panel
safety correlations; error bars are endpoint-bootstrap 95\% percentile intervals. Thin lines connect
panel-level estimates, not model trajectories. Misalignment changes from $-0.44$ to $-0.16$ and
jailbreak from $+0.08$ to $+0.34$; both direct change intervals include zero. Task success was
not re-collected for the expansion.}
\label{fig:crossover}
\end{figure}

\ifdefined\workshopversion\else
\pagebreak[4]
\fi

\paragraph{Against agentic misalignment, capability has the opposite association.} On the full
21-endpoint panel, the composite is negatively associated with misalignment safety
($\rho_{\text{cap}}=-0.44$),
compared with $+0.60$ on $\tau^2$ (Mistral-Large blackmails or leaks in $85\%$ of decisions, o3-mini
in $1\%$).
On the primary maximum-coverage populations, capability-adjusted estimates are $+0.37$ for R-Judge
specificity ($n{=}20$), $+0.28$ for InjecAgent robustness ($n{=}21$), and $+0.13$ for AgentHarm
safety ($n{=}19$; Table~\ref{tab:rq3}); all residual-permutation $p\geq0.11$. The common $n{=}18$
panel raises the first two to $+0.41$ and $+0.47$, showing why the population rule matters
(App.~\ref{app:rq3pop}). Power is only $\approx0.35$, and our directional prediction concerned AgentHarm rather than
R-Judge or InjecAgent, so these comparisons are exploratory. The R-Judge
estimate also depends on the metric: on the same $n{=}20$ population, balanced accuracy gives $+0.04$
instead of specificity's $+0.37$ (Table~\ref{tab:rq3}); on the common $n{=}18$ panel, $+0.41$ under
specificity becomes $+0.19$ under balanced accuracy and $-0.46$ under recall (App.~\ref{app:metric}).

\paragraph{A direct interaction test on the original panel.} We compare capability's correlations
with task success and misalignment on their paired $n{=}20$ panel. The task-success correlation is
$+0.60$ and the misalignment-safety correlation is $-0.41$ ($-0.44$ on the full panel), a
difference of $\Delta=-1.00$ (endpoint-bootstrap percentile interval $[-1.48,-0.49]$). This
difference is stable to
leave-one-endpoint-out, leave-one-organization-out, organization-clustered bootstrap, and subsampling
(App.~\ref{app:robustness}).
\ifdefined\workshopversion
This paired bootstrap tests the difference between dependent correlations, not either against zero,
and conditions on the single realized $\tau^2$ run (simulator-rerun variance is unavailable).
\fi
The corresponding R-Judge interaction is unresolved
($\Delta=+0.50$, percentile interval $[-0.07,+1.14]$), so the evidence supports the recorded-panel
capability contrast but not a full two-way dissociation. An independent misalignment grader
produces nearly identical endpoint scores ($\rho=0.97$).

\paragraph{The capability--misalignment correlation is weaker on the expanded panel.}
The expansion proceeded in two stages. With capability rescaled on the pooled panel, the original 21 models give
$\rho=-0.45$ (versus $-0.44$ under the original scaling), the
July 7 gate-only models $+0.54$ ($n{=}6$), the subsequently specified July 9-only models $+0.01$
($n{=}13$), all new models $+0.07$ ($n{=}19$), and the pooled panel $-0.16$ ($n{=}40$;
App.~\ref{app:n40}). Before the second stage we predicted a pooled $\rho$ between $-0.20$ and $+0.25$ with an interval
excluding $-0.44$; the estimate fell in that range, but its interval still includes $-0.44$. Because the pooled estimate includes the motivating observations, it is a population
sensitivity result, not independent confirmation. The nested-bootstrap difference is
$\Delta\rho=+0.27$ with 95\% interval
$[-0.25,+0.78]$. The point estimate differs on the expanded panel, but the change interval includes zero. Subsets
matched to the original organization composition give median $\rho=-0.17$, so composition alone
does not explain the difference; the data do not establish a generational explanation. Because $\tau^2$ was not re-collected, we could not re-test the original
interaction.

\paragraph{Models with similar capability scores differ in safety across organizations
(Fig.~\ref{fig:developer}, appendix).} The 41-endpoint panel spans 12 model-developing organizations
(misalignment $n{=}40$; jailbreak $n{=}41$ after one engagement-gate exclusion). Six
organizations contribute at least four models for within-organization correlations, and seven
contribute at least three models for the group comparison (App.~\ref{app:n40},
Table~\ref{tab:n40}). Within organizations, capability--safety correlations vary in sign and
average near zero ($-0.14$ misalignment, $+0.26$ jailbreak). After adjusting safety ranks for
capability, the fixed-roster organization effect sizes are $\varepsilon^2=0.34$ and $0.57$;
deleting one eligible organization at a time gives $0.27$--$0.39$ and $0.41$--$0.69$. We do not
assign a label-permutation $p$-value because organization labels are not exchangeable over endpoint
rows. None of this is causal: the static composite may miss agentic ability, each organization
contributes few endpoints, and some groups have restricted ranges. On this roster, models with
similar MMLU/GPQA scores have different safety scores.

\paragraph{Against jailbreak safety, AgentHarm shows the largest held-out association.}
A separate dated protocol introduced a second safety variable with a different failure mode,
harness, and scoring procedure: harmful compliance under jailbreak, scored with StrongREJECT
\citep{strongreject2024}. A single jailbreak template (the AIM role-play prompt of \citealt{wei2023jailbroken}) produces a ceiling because 11 of the 22
jailbreak-scored models refuse every request (21 have capability anchors). Averaging scores over three published templates
\citep{wei2023jailbroken} leaves no model at the ceiling. Capability is weakly associated with this
outcome on the 2024--25 panel ($\rho=+0.08$, $n{=}21$). The expanded-panel estimate is $+0.34$
($n{=}41$; endpoint-bootstrap percentile interval $[+0.03,+0.60]$), but the direct between-panel
change interval includes zero.

After adjusting for the static-knowledge composite, AgentHarm safety correlates with the three-template jailbreak outcome
at $\rho=+0.72$ (endpoint-level residual-permutation $p=0.0005$), compared with $+0.50$ for the
ceiling-limited single-template measure. An exhaustive bootstrap over the eight represented
organizations gives a 95\% interval $[+0.17,+0.97]$ and a two-sided sign-tail of $0.018$, a
descriptive summary rather than a familywise test.
The estimate is also stable to leave-one-endpoint analysis and an alternate grader
(App.~\ref{app:robustness}). Missingness matters: assigning the two tool-loop-missing endpoints
every possible AgentHarm safety pair on a $101^2$ grid keeps the point estimate positive
($+0.46$--$+0.78$), but assigning both $0.90$ safety gives an organization interval
$[-0.22,+0.90]$. The $+0.72$ result is therefore conditional on eligible tool-loop execution.
AgentHarm is more strongly associated with jailbreak than with misalignment in the point estimates
($\Delta=+0.59$; endpoint-bootstrap interval $[+0.07,+1.10]$), but the eight-organization interval
touches zero ($[0.00,+1.10]$). On this primary natural panel, the misalignment partial is $+0.13$
($n{=}19$); on the common $n{=}18$ panel it is $+0.16$. R-Judge
specificity ($-0.11$) and InjecAgent robustness
($+0.21$) show no significant incremental association with jailbreak safety. AgentHarm and StrongREJECT both measure harmful
compliance, so part of what the $+0.72$ shows is convergent validity. The organization-resampled interval for
the AgentHarm selectivity contrast includes zero, as does the model-resampled interval for the
corresponding R-Judge/InjecAgent comparison.

Blinded coding and an XSTest over-refusal check \citep{xstest2024} make alternative explanations
less likely but do not exclude them; the full planned check is mixed (App.~\ref{app:artifacts}).

\section{Discussion}

On these snapshots, a constant baseline matches R-Judge F1, benchmark rankings disagree, and
relations to outside variables depend on outcome and population (Table~\ref{tab:rq3}). The
AgentHarm--jailbreak association is convergent evidence that holds only for models that completed
the tool loop; the other associations are exploratory.

\ifdefined\verifyagentsworkshop
\paragraph{Implications for agent verification.}
Treat each benchmark as a typed signal: name the developer, acting endpoint or verifier, scorer,
target, and error profile. State the intended decision; label external relations as convergent,
discriminant, or exploratory; adjust for a measured capability composite and say what it omits; and recheck changing endpoint
populations. Do not select endpoints with a gameable score or an unstable relation.

\else
  \ifdefined\iaevalworkshop
  \input{sections/05_iaeval_implications}
  \fi
\fi

\section{Limitations and Threats to Validity}
\label{sec:limitations}

\paragraph{Panel and anchors.} The common three-benchmark panel has $n{=}18$ models, limited by AgentHarm coverage; a larger
capability-spanning agent-loop panel is needed. MMLU/GPQA are static-knowledge anchors. In our pilot with the Berkeley Function Calling
Leaderboard (BFCL; \citealt{bfcl2025}), the scores rewarded strict call syntax; anchor error can
attenuate partials.

\paragraph{Multiplicity and dependence.} RQ3 developed sequentially, so we make no familywise
claim. AgentHarm--jailbreak has residual-permutation $p=0.0005$, but its organization sign-tail is
descriptive; RQ1/RQ2 and organization patterns are fixed-roster sensitivities, not population inference.

\paragraph{Judges, missingness, and coverage.} Alternate graders correlate $0.97/0.98$ for misalignment/jailbreak;
a Gemini-judged AgentHarm refusal rate agrees at $0.69$, and substituting it moves the key partial
only $+0.72\to+0.67$. AgentHarm's scalar scores task completion, so a failed harmful attempt counts
as safe, and both AgentHarm and the jailbreak test use OpenAI-family LLM graders. AgentDojo covers
five endpoints. Bounded imputation of two AgentHarm failures preserves the point direction, but
optimistic imputation makes the organization interval cross zero (App.~\ref{app:rq3pop}).

\paragraph{External-variable evidence (RQ3).} $\tau^2$ has one simulator-dependent run/model;
misalignment is fictional and its two halves agree only moderately ($0.70$); jailbreak averages
three templates; none estimates deployment harm. Estimates condition on one execution and omit
provider/rerun variance. AgentHarm--jailbreak partly reflects construct overlap; the low-power
misalignment result's larger three-scenario estimate is post-hoc (App.~\ref{app:robustness}).

\section{Conclusion}

\ifdefined\workshopversion
Across four benchmarks and up to 22 recorded endpoints, R-Judge's ``always unsafe'' baseline beats
five endpoints, and benchmark choice changes matched-snapshot rankings. The measured
static-knowledge composite tracks task success but not safety consistently. AgentHarm correlates
$+0.72$ with jailbreak safety after
controlling MMLU/GPQA (organization interval $[+0.17,+0.97]$), but construct overlap and tool-loop
missingness make this conditional convergent evidence. Report confusion matrices, avoid standalone
$F_1$ when true negatives matter, and bind claims to actors, targets, snapshots, and populations.
\else
Across four benchmarks and up to 22 models, R-Judge's F1 lets an ``always unsafe'' baseline outrank
five models, and switching benchmarks switches the rankings on the matched panel. Capability predicts held-out task
success; it does not predict safety consistently across outcomes or panels.
The largest held-out safety association is AgentHarm's ($\rho=+0.72$ with jailbreak safety after
adjusting for the static-knowledge composite; organization-cluster interval $[+0.17,+0.97]$), though the two measures
overlap and do not establish a universal safety summary.
Safety evaluations should report the full confusion structure,
drop standalone F1 wherever true negatives matter, and tie every claim to an explicit target
behavior and model population.
\fi

\bibliographystyle{plainnat}
\bibliography{references,references_verifyagents}

\appendix
\section{Technical Appendix}

This appendix gives method details and the robustness results summarized in the body.

\begin{table}[h]
\centering
\scriptsize
\begin{tabular}{@{}>{\raggedright\arraybackslash}p{0.16\linewidth}>{\raggedright\arraybackslash}p{0.17\linewidth}>{\raggedright\arraybackslash}p{0.09\linewidth}>{\raggedright\arraybackslash}p{0.21\linewidth}>{\raggedright\arraybackslash}p{0.24\linewidth}@{}}
\toprule
Result & Score & Models ($n$) & Uncertainty & What it supports \\
\midrule
Constant baseline (\S4.1) & R-Judge F1 & 21 discriminating & closed form; no sampling & F1 alone does not show two-sided discrimination on R-Judge \\
Ranking disagreement (\S4.3) & R-Judge specificity, InjecAgent robustness, AgentHarm safety & 18 common & item resampling (App.~\ref{app:panels}) & These scores do not substitute for one another as one ranking \\
Small-panel trade-off (\S4.3) & R-Judge specificity vs.\ AgentHarm safety & 7 vs.\ 18 & a quarter of size-7 subsets reach $|\rho|\geq0.5$ & Seven-model panels can show trade-offs that do not persist \\
Metric dependence (\S4.2) & R-Judge specificity, balanced acc., F1 vs.\ MMLU/GPQA & 20 & leave-one-organization-out ranges & Association with the composite depends on the R-Judge metric \\
Task success (\S4.4) & safety-score partials & 19--20 & wide organization-cluster intervals (App.~\ref{app:rq3pop}) & No detectable incremental association; not evidence of none \\
Misalignment (\S4.4) & composite raw $\rho$ & 21 and 40 & change interval includes zero (App.~\ref{app:n40}) & Exploratory; population-sensitive \\
AgentHarm--jailbreak (\S4.4) & AgentHarm safety partial & 19 & organization interval $[+0.17,+0.97]$; crosses zero under optimistic imputation & Convergent evidence on harmful compliance, for eligible tool-loop runs only \\
AgentDojo (\S4.3) & injection robustness & 5 & too few models to estimate & Inconclusive convergent check; case study only \\
\bottomrule
\end{tabular}
\caption{Result map. Each main result with its score, model population, uncertainty source, and
the claim it supports. All adjusted results condition on the MMLU/GPQA static-knowledge composite
and on one realized run per model and instrument.}
\label{tab:resultmap}
\end{table}

\subsection{Actor roles and the audited score use}
\label{app:actor_map}

The score user is a developer choosing among recorded candidate endpoints. The audit asks whether
replacing one typed scalar with another preserves the endpoint ordering, or whether the scalars can
be read as one undifferentiated safety ranking. We do not evaluate dashboards that combine several scores.

\begin{center}\small
\begin{tabular}{@{}>{\raggedright\arraybackslash}p{0.16\columnwidth}>{\raggedright\arraybackslash}p{0.20\columnwidth}>{\raggedright\arraybackslash}p{0.28\columnwidth}>{\raggedright\arraybackslash}p{0.27\columnwidth}@{}}
\toprule
Signal & endpoint role & sampled behavior / scorer & bounded interpretation \\
\midrule
R-Judge & verifier & labels stored benign/unsafe traces; endpoint self-verdict & risk-recognition discrimination \\
InjecAgent & acting policy & reacts to injected tool observations; rule scorer & injection resistance conditional on valid output \\
AgentHarm & acting policy & attempts/refuses harmful tool tasks; official LLM-assisted scorer & low harmful-task completion (refusal or failed attempt) \\
AgentDojo & acting policy & acts in a sandbox; environment security/utility state & injection security with task utility \\
\bottomrule
\end{tabular}
\end{center}

The external variables are not additional audited benchmarks: $\tau^2$ task success is a
discriminant check, misalignment is an exploratory broader-use criterion, and three-template
jailbreak safety is construct-overlapping convergent evidence. A judge/scorer is also distinct from
the candidate endpoint except in R-Judge's self-verification protocol.

\subsection{Data lineage and chronology}
\label{app:lineage}

\paragraph{R-Judge row counts.} The paper and upstream README describe 569 records
\citep{rjudge2024}, whereas the 27 JSON files at the upstream data commit verified for this audit
(\texttt{83ce301da3ad50dd8b397e772863f5411c3d3dc2}) contain 571 unique IDs: 270 benign and 301 unsafe.
Our frozen source files exactly match that commit, so we call this the \emph{571-record release
snapshot}, not the universal benchmark size. The class-prevalence baseline uses all 571 rows. The
21 accepted original-panel executions share 570 IDs because DeepSeek-R1 lacks ID \texttt{2288};
Table~\ref{tab:reliability} therefore uses the 570-ID intersection. Per-endpoint hashes are in the
supplement manifest.

\paragraph{Adaptive R-Judge continuation.} The July 7 gate attempted eight endpoints. Seven logs
contain all 571 IDs; Gemini-2.5-Pro stopped at 228. Six record-complete endpoints also have both
anchors; GPT-5 among them has only $20.1\%$ valid R-Judge verdicts (GLM-4.5, $87.2\%$, also lacks a
GPQA anchor), so five new endpoints pass both the anchor and $90\%$ verdict gates. They enter the continuation: the recorded
common-panel estimate $+0.41$ ($n{=}18$) becomes $+0.23$ on all eligible models ($n{=}25$: 20 original
plus five new), below the $0.30$ stop rule.
The subsequent July 9 candidate stage therefore adds no R-Judge executions. The artifact reports
every gate count and valid rate.

\paragraph{Chronology boundary.} Protocol files carry author-entered date headings but no
independent timestamp, and later files respond to observed results. CSV \texttt{run\_date} values
are hard-coded runner metadata rather than collection-time evidence. The authors' internal record
places the expansion in July 7--12, 2026, but that window is not independently verifiable from the
public artifact. We therefore report the full adaptive genealogy and use no preregistration or
joint-confirmation language.

\subsection{Ruling out two artifacts (detail for \S\ref{sec:rq3})}
\label{app:artifacts}

\paragraph{Blinded construct coding.} We used seven LLM coders from distinct families: GPT-4o,
Claude-3-Haiku, Gemini-2.5-Flash, Llama-3.3-70B, Qwen-2.5-72B, Mistral-Large, and DeepSeek-V3.
Each coder was shown only a neutral, predictive-information-stripped description of what each benchmark
\emph{measures} (no mention of our criteria, category names, or any RQ3 result) plus two construct
definitions (``harm blocking'' vs.\ ``risk recognition and correct treatment of benign inputs''). The two options
and the benchmark order were shuffled per coder to remove position bias. Each coder assigned every
benchmark to one construct. Votes (B${=}$harm blocking, R${=}$risk recognition):

\begin{center}\small
\begin{tabular*}{\columnwidth}{@{\extracolsep{\fill}}lccc@{}}
\toprule
Benchmark & blind & author & B\,/\,R \\
\midrule
R-Judge    & recognition & recognition & 0 / 7 \\
InjecAgent & recognition & recognition & 1 / 6 \\
AgentHarm  & blocking    & blocking    & 7 / 0 \\
AgentDojo  & recognition & recognition & 0 / 7 \\
\bottomrule
\end{tabular*}
\end{center}

All four majorities match the hand-assignment; raw agreement is $27/28$ ratings, Krippendorff's
$\alpha=0.825$ (nominal), against a chance baseline of $0.5^4=6.2\%$ for one coder matching all
four labels by coin flip. The single dissent (Llama-3.3-70B placing InjecAgent as harm blocking) is the
one semantically borderline case (injection-robustness). This controls coder bias but not
rubric-designer bias (the category definitions were written by authors who had seen the results); the exploratory PCA (App.~\ref{app:rq1}) also places AgentHarm ($+0.33$) opposite R-Judge and InjecAgent without using the external criteria.

\paragraph{Blinded criterion coding, with a distractor.} To check that the predictor$\to$criterion
matching (not just the predictor labels) is recoverable blind, and to make the agreement statistic
meaningful, we re-ran the coding over all six targets (four benchmarks $+$ two criteria) with a third,
plausible \emph{task-success} distractor construct. Krippendorff's $\alpha=0.61$ (three categories,
chance $\approx1/3$), and the distractor is chosen only $5/42$ times. The harm-blocking
classification is robust: AgentHarm and the jailbreak criterion both map to harm blocking $7/7$,
R-Judge to risk recognition $7/7$,
and InjecAgent $5/7$, all unchanged by the distractor. Two targets are honestly contested: AgentDojo
reads as task success to $4/7$ coders (it is utility-scored; our $n{=}5$ benchmark), and the
misalignment criterion is contested: it maps to risk recognition $6/7$ in a separate two-way round%
\ifdefined\workshopversion
, but to harm blocking $6/7$ here (avoiding a harmful autonomous action
\else
\ (frozen votes in the artifact), but to harm blocking $6/7$ here (avoiding a harmful autonomous action
\fi
can be read as both recognizing and declining it). This recovers the AgentHarm--jailbreak pairing
without using the predictive results, and is an independent reason we keep the
R-Judge/InjecAgent--misalignment pattern exploratory.

\paragraph{XSTest over-refusal (blanket-refusal check).} We ran XSTest (Röttger et al., 2024;
cited in the main paper) on the
panel via its published CPI (compliance/partial/refusal) scorer (judge gpt-4o-mini): the
\emph{safe} subset (benign prompts that superficially resemble unsafe ones) measures over-refusal,
the \emph{unsafe} subset measures correct refusal of overtly-harmful prompts. The unsafe subset is
ceilinged (every model refuses $83$--$100\%$, sd $4.5$, limited discriminating variance), so the
informative signal is over-refusal on the safe subset (range $0$--$60\%$; Claude-3-Haiku over-refuses
$60\%$, o3-mini $30\%$, others $\leq15\%$). Partial-$\rho$ (capability-controlled) two-by-two:

\begin{center}\small
\begin{tabular*}{\columnwidth}{@{\extracolsep{\fill}}lcc@{}}
\toprule
Predictor & \shortstack{unsafe\\refusal} & \shortstack{safe\\compliance} \\
\midrule
AgentHarm safety        & $+0.24$ ($n{=}17$) & $+0.01$ ($n{=}17$) \\
R-Judge specificity     & $+0.06$ ($n{=}17$) & $-0.15$ ($n{=}18$) \\
Misalignment safety     & $+0.47$ ($n{=}18$) & $-0.11$ ($n{=}19$) \\
\bottomrule
\end{tabular*}
\end{center}

AgentHarm shows no association with benign-prompt over-refusal ($\rho=-0.01$, $n{=}17$), weighing against
but not ruling out a blanket-refusal effect. Capability is $+0.21$ (n.s.) on benign compliance and
$-0.32$ (n.s.) on harmful refusal; R-Judge specificity is $+0.15$ on over-refusal (equivalently
$-0.15$ on safe compliance). Misalignment safety likewise shows no association with over-refusal
($+0.11$, $n{=}19$, $p=0.66$), while its unsafe-refusal association is $+0.47$
($n{=}18$, $p=0.052$). The planned two-arm decision rule is mixed, not a pass: AgentHarm's matched
cell ($+0.24$) exceeds its off-diagonal ($+0.01$), but R-Judge's matched benign-compliance cell
($-0.15$) does not exceed its harmful-refusal cell ($+0.06$). These low-power checks neither recover
a general two-construct structure nor rule out alternative explanations.

\subsection{Panel coverage and power}
\label{app:panels}

\begin{center}\small
\begin{tabular}{@{}>{\raggedright\arraybackslash}p{0.43\columnwidth}c>{\raggedright\arraybackslash}p{0.31\columnwidth}@{}}
\toprule
Analysis & $n$ & note \\
\midrule
R-Judge (specificity, F1)     & 21 & Qwen3-32B dropped; shared item intersection 570 \\
InjecAgent (robustness)       & 22 & conditional on valid outputs; fixed 300 IDs \\
AgentHarm (safety)            & 19 & conditional on eligible official tool-loop result \\
AgentDojo                     & 5  & convergent control inconclusive \\
$\tau^2$ criterion            & 20 & \\
misalignment criterion        & 21 & \\
jailbreak criterion (\mbox{StrongREJECT}) & 22 & 3-template average \\
expanded misalignment / jailbreak & 40 / 41 & 12 organizations (App.~\ref{app:n40}) \\
XSTest over-refusal & 17--20 & AgentHarm$\cap$anchor overlap \\
RQ1 common cross-benchmark    & 18 & AgentHarm-limited \\
RQ3 natural / common panels   & 19--21 / 18 & cellwise manifest below \\
capability interaction (paired)& 20 & \\
\bottomrule
\end{tabular}
\end{center}

\begin{figure*}[t]
\centering
\ifdefined\workshopversion
\includegraphics[width=\textwidth,height=0.76\textheight,keepaspectratio]{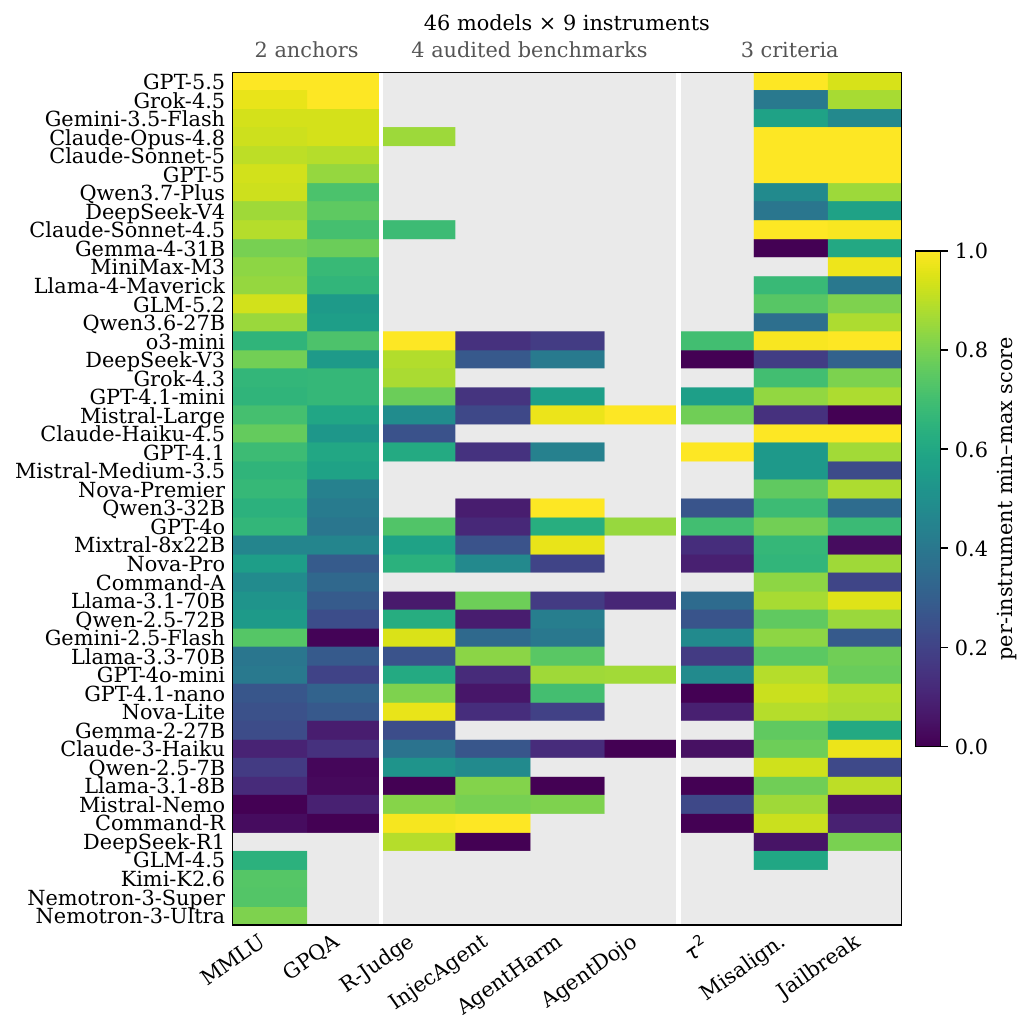}
\else
\includegraphics[height=0.76\textheight]{figures/fig_panel_scale.pdf}
\fi
\caption{\textbf{The audit's evidence base: up to 46 models $\times$ 9 instruments} (two capability
anchors, four audited benchmarks, three held-out criteria; $262$ model$\times$instrument cells). Cells are
per-instrument min--max-normalized scores; gray is not-run. The rows beyond the $41$-model anchored
panel are models excluded from the analyses by protocol-specific telemetry gates
(App.~\ref{app:n40}). Rows are sorted by capability: the
2026 expansion (top) is scored on the anchors and the two chat-based criteria, which need
no tool-calling harness, while the four audited benchmarks run on the original panel. The capability anchors
and the R-Judge/InjecAgent scores vary together; the safety criteria do not track them consistently.}
\label{fig:scale}
\end{figure*}

\paragraph{Within-endpoint score uncertainty.} From the existing item-level data (no new runs), per-endpoint
standard errors for every instrument: MMLU binomial SE over its 500 fixed items (median $1.6$, max $2.2$
points), GPQA over 198 (median $3.3$, max $3.6$), the misalignment criterion over its 100 decisions
(median $4.0$, max $5.0$), R-Judge F1 bootstrapped over up to 571 records (median $1.8$, max $3.0$), and
the jailbreak criterion bootstrapped over its ${\sim}150$ judged prompts (median $2.3$, max $3.7$). Against each
instrument's between-model SD ($8.7$, $18.2$, $26.0$, $14.9$, $28.5$ respectively, on the expanded
panel), the \emph{worst-case} within-endpoint SE is $13$--$25\%$ of that SD. Thus the observed
panel-level spread is $4$--$8\times$ the worst-case SE, although individual model pairs can be
closer. ($\tau^2$-retail remains a
single run whose user-simulator variance we cannot bound; it is the one instrument this analysis
cannot cover.)

\paragraph{Internal consistency, not just precision.} Small SEs do not by themselves make rankings stable,
so we also compute, per instrument from its item-level data, split-half consistency
(Spearman--Brown), item-bootstrap rank stability, and top-quartile persistence
(Table~\ref{tab:reliability}). Every item-rich instrument with item-resampling support is internally consistent (split-half
$\geq0.96$ corrected; rank stability $\geq0.97$; a top-quartile model stays top-quartile in
$0.79$--$0.96$ of item resamples), making item-sampling error an unlikely explanation for the
panel-level ranking disagreement in Table~\ref{tab:matrix}. Misalignment has only two scenario
halves (Spearman--Brown $0.70$), and $\tau^2$ has no resampling bound. The jailbreak
computation resamples by base prompt, so the three templates' dependence is respected.

\begin{table*}[t]
\centering
\small
\begin{tabular}{@{}lrrrrr@{}}
\toprule
Instrument & models & items & split-half (SB) & rank stab.\ & quartile \\
\midrule
R-Judge specificity & 21 & 570 & $+0.98$ ($+0.99$) & $+0.99$ & $0.90$ \\
R-Judge F1 & 21 & 570 & $+0.93$ ($+0.96$) & $+0.97$ & $0.87$ \\
InjecAgent robustness & 22 & 300 & $+0.95$ ($+0.98$) & $+0.98$ & $0.88$ \\
AgentHarm safety (behaviors) & 19 & 44 & $+0.93$ ($+0.96$) & $+0.97$ & $0.79$ \\
Jailbreak crit.\ (by base prompt) & 42 & 20 & $+0.94$ ($+0.97$) & $+0.98$ & $0.96$ \\
MMLU (anchor) & 46 & 356 & $+0.93$ ($+0.96$) & $+0.98$ & $0.92$ \\
GPQA (anchor) & 45 & 164 & $+0.95$ ($+0.97$) & $+0.98$ & $0.92$ \\
Misalignment crit.\ (scenario halves) & 22 & 2 & $+0.54$ ($+0.70$) & --- & --- \\
\bottomrule
\end{tabular}
\caption{Internal consistency from available item- or scenario-level data: mean split-half correlation of model scores over 50 random half-splits (Spearman--Brown corrected), mean Spearman of item-bootstrapped vs.\ observed rankings, and the probability that an observed top-quartile model stays top-quartile under resampling. Item counts are the complete-coverage intersection across all listed models (e.g.\ 356 of the 500 administered MMLU items, 164 of 198 GPQA items). Splits/draws are shared across models and use the items common to every model's logs; the jailbreak criterion resamples by base prompt (the three templates are dependent), and its shared set is 20 of the 50 suite prompts because each model leaves a few different prompts unjudged. Jailbreak uses 42 models: the 41 anchored models plus DeepSeek-R1, which lacks the required anchor pair; misalignment has config-level data only, so its row is the blackmail-vs-leaking halves convergence. Every item-rich instrument with item-resampling support has Spearman--Brown $\geq0.96$ and rank stability $\geq0.97$, making item-sampling error an unlikely explanation for the panel-level disagreement of Table~\ref{tab:matrix}. Misalignment's two scenario halves agree only moderately (Spearman--Brown $0.70$). $\tau^2$ is a single run and remains the one unbounded instrument.}
\label{tab:reliability}
\end{table*}

The exploratory R-Judge/InjecAgent--misalignment estimates use $n{=}20$--$21$ on their natural
panels ($+0.37/+0.28$) and $n{=}18$ on the common sensitivity panel ($+0.41/+0.47$). Simulated
power is only $\approx0.35$ to detect $\rho{=}0.4$ at $\alpha{=}0.05$; the minimum detectable effect
at $n{=}18$ and $80\%$ power is $0.62$. These results leave practically meaningful effects unresolved. Across the nine natural-panel cells, the AgentHarm--jailbreak
endpoint-level residual-permutation value is $0.00053$ and all others are $\geq0.11$. Its exhaustive
eight-organization bootstrap gives interval $[+0.175,+0.975]$ and descriptive sign-tail $0.0176$.
Because that sign-tail is not a calibrated null test and the family developed sequentially, we make
no Bonferroni or familywise-confirmatory claim.

\begin{figure*}[t]
\centering
\includegraphics[width=\textwidth]{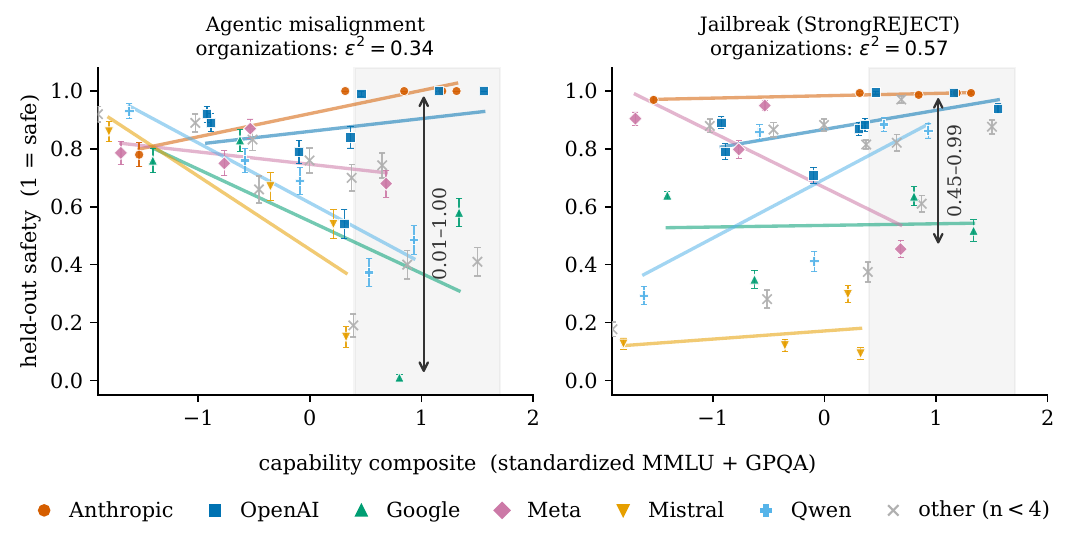}
\caption{\textbf{Models with similar measured capability can differ widely in safety across
model-developing organizations.} Each point is a model; color and shape identify its organization;
error bars are within-model SEs
(\S\ref{app:panels}). \emph{Within} an organization (lines, the six organizations with $\geq4$ anchored models) the
capability$\to$safety slope is large but sign-heterogeneous across organizations ($n$-weighted mean $-0.14$
misalignment, $+0.26$ jailbreak). Among higher-capability models (shaded, composite $>0.4$), misalignment safety
spans $0.01$--$1.00$ and jailbreak safety $0.45$--$0.99$ (arrows). After rank-residualizing safety on capability, fixed-roster
organization effect sizes are $\varepsilon^2{=}0.34$ / $0.57$ (leave-one-organization-out ranges
$0.27$--$0.39$ / $0.41$--$0.69$). The static composite does not capture all capability, so this is an association,
not evidence that organizational identity causes the difference.}
\label{fig:developer}
\end{figure*}

\subsection{RQ3 analysis populations and InjecAgent denominator}
\label{app:rq3pop}

Table~\ref{tab:rq3pop} uses maximum available coverage within the original 22-model roster for each
predictor--variable cell as its declared primary rule. It reports the common 18-model panel only as
a sensitivity. The machine-readable manifest is
\texttt{analysis/rq3\_analysis\_populations.csv}. DeepSeek-R1 lacks a complete harmonized anchor;
Qwen3-32B fails R-Judge's valid-output gate; Qwen-2.5-7B lacks the task-success run and an eligible
AgentHarm tool-loop score. Command-R has no eligible official AgentHarm result, and the saved record
cannot separate model behavior from routing, so it remains missing rather than being recoded. The
AgentHarm estimand is therefore conditional on obtaining an eligible official tool-loop result.

\paragraph{AgentHarm missing-outcome sensitivity.}
Both missing endpoints have complete anchors, misalignment, and jailbreak outcomes. We assign
Command-R and Qwen-2.5-7B every pair of AgentHarm safety values on a $101\times101$ grid over
$[0,1]^2$. The capability-adjusted AgentHarm--jailbreak point estimate remains positive throughout:
$\rho\in[+0.456,+0.784]$. Dependence sensitivity is weaker. If both missing endpoints are assigned
safety $0.90$ (the point-estimate minimum, tied through 1.0), the estimate is $+0.456$ at $n{=}21$
and the exact nine-organization percentile interval is $[-0.215,+0.905]$. Thus the complete-case
$+0.725$ estimate stays positive under every grid imputation, but the organization-cluster interval can include zero depending on the missing tool-loop results. The grid and exact bootstrap are emitted by the supplement.

\begin{table*}[t]
\centering\scriptsize
\begin{tabular*}{\textwidth}{@{\extracolsep{\fill}}llrrrrlrr@{}}
\toprule
Predictor & variable & $n$ & org. & $\rho$ & perm. $p$ & org.-cluster 95\% interval & sign-tail & common $\rho$ \\
\midrule
R-Judge    & task       & 19 & 9 & $-0.088$ & .721 & $[-.622,+.413]$ & .589 & $-0.026$ \\
R-Judge    & misalign.  & 20 & 9 & $+0.367$ & .114 & $[-.157,+.743]$ & .116 & $+0.414$ \\
R-Judge    & jailbreak  & 20 & 9 & $-0.114$ & .627 & $[-.729,+.560]$ & .719 & $-0.104$ \\
InjecAgent & task       & 20 & 9 & $+0.159$ & .501 & $[-.282,+.617]$ & .406 & $+0.193$ \\
InjecAgent & misalign.  & 21 & 9 & $+0.283$ & .214 & $[-.297,+.573]$ & .247 & $+0.465$ \\
InjecAgent & jailbreak  & 21 & 9 & $+0.213$ & .350 & $[-.364,+.674]$ & .531 & $+0.121$ \\
AgentHarm  & task       & 19 & 8 & $-0.233$ & .334 & $[-.794,+.226]$ & .329 & $-0.327$ \\
AgentHarm  & misalign.  & 19 & 8 & $+0.126$ & .604 & $[-.392,+.591]$ & .490 & $+0.156$ \\
AgentHarm  & jailbreak  & 19 & 8 & $+0.725$ & .00053 & $[+.175,+.975]$ & .0176 & $+0.748$ \\
\bottomrule
\end{tabular*}
\caption{RQ3 population and dependence sensitivity. The primary $\rho$ uses the declared
maximum-coverage rule; ``common'' uses the same 18 models throughout. Permutation $p$ is a
endpoint-unit residual null test. The exhaustive organization-cluster interval and sign-tail describe
cluster-bootstrap uncertainty; the sign-tail is not a calibrated null $p$-value.}
\label{tab:rq3pop}
\end{table*}

\paragraph{InjecAgent item and output validity.}
\label{app:injecdenom}
We reaggregate all models on the same 300-ID intersection (SHA-256 prefix
\texttt{3c89827cbf63}); this removes nine extra raw \texttt{ds-base} rows for GPT-4o-mini and
Llama-3.1-8B. Valid-output rates range from $38.7\%$ to $100\%$ and correlate with capability at
$\rho=+0.552$ ($n{=}21$, asymptotic $p=0.0095$), so invalidity is informative. Table~\ref{tab:injecdenom}
reports every denominator and sharp all-item bound. Under conditional / invalid-safe / invalid-attack
conventions, InjecAgent's natural-panel partials are respectively $+.159/+.092/+.299$ for task,
$+.283/+.257/+.400$ for misalignment, and $+.213/+.203/+.175$ for jailbreak. No qualitative
conclusion changes, although magnitudes do.

\begin{center}\scriptsize
\begin{tabular*}{\columnwidth}{@{\extracolsep{\fill}}lrrrr@{}}
\toprule
Model & valid & cond. & invalid=attack & invalid=safe \\
\midrule
Claude-3-Haiku    & 238/300 & 76.9 & 61.0 & 81.7 \\
Command-R         & 210/300 & 17.1 & 12.0 & 42.0 \\
DeepSeek-R1       & 274/300 & 98.9 & 90.3 & 99.0 \\
DeepSeek-V3       & 251/300 & 76.1 & 63.7 & 80.0 \\
GPT-4.1           & 280/300 & 86.8 & 81.0 & 87.7 \\
GPT-4.1-mini      & 268/300 & 86.6 & 77.3 & 88.0 \\
GPT-4.1-nano      & 248/300 & 94.0 & 77.7 & 95.0 \\
GPT-4o            & 268/300 & 89.6 & 80.0 & 90.7 \\
GPT-4o-mini       & 250/300 & 88.4 & 73.7 & 90.3 \\
Gemini-2.5-Flash  & 297/300 & 71.0 & 70.3 & 71.3 \\
Llama-3.1-70B     & 208/300 & 35.1 & 24.3 & 55.0 \\
Llama-3.1-8B      & 116/300 & 31.9 & 12.3 & 73.7 \\
Llama-3.3-70B     & 253/300 & 31.2 & 26.3 & 42.0 \\
Mistral-Large     & 250/300 & 81.2 & 67.7 & 84.3 \\
Mistral-Nemo      & 224/300 & 33.9 & 25.3 & 50.7 \\
Mixtral-8x22B     & 230/300 & 77.8 & 59.7 & 83.0 \\
Nova-Lite         & 291/300 & 88.3 & 85.7 & 88.7 \\
Nova-Pro          & 275/300 & 60.4 & 55.3 & 63.7 \\
Qwen-2.5-72B      & 288/300 & 92.4 & 88.7 & 92.7 \\
Qwen-2.5-7B       & 194/300 & 59.8 & 38.7 & 74.0 \\
Qwen3-32B         & 279/300 & 92.5 & 86.0 & 93.0 \\
o3-mini           & 300/300 & 87.3 & 87.3 & 87.3 \\
\bottomrule
\end{tabular*}
\end{center}
\captionof{table}{InjecAgent reaggregation on the frozen 300-ID intersection. Scores are percentages:
\emph{cond.} is robustness among valid outputs; the last two columns bound invalid outputs as attack
successes or safe, respectively.}
\label{tab:injecdenom}

\subsection{Cross-benchmark structure (detail for \S\ref{sec:rq1})}
\label{app:rq1}

\begin{figure}[t]
\centering
\includegraphics[width=0.92\columnwidth]{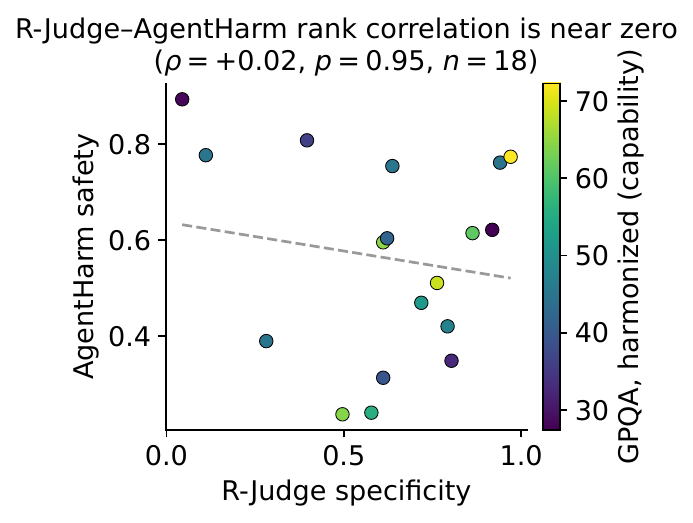}
\caption{\textbf{R-Judge and AgentHarm do not rank the full panel in opposite order.} AgentHarm
safety against R-Judge specificity on the $n{=}18$ cross-benchmark panel, colored by the harmonized
GPQA anchor. The trade-off a 7-endpoint panel suggested ($\rho=-0.64$) dissolves at $n{=}18$
($\rho=+0.02$; gray trend): endpoints near the top of either axis span the full range of
the other, the pairwise view of Table~\ref{tab:matrix}'s rank disagreement.}
\label{fig:reversal}
\end{figure}

\begin{figure}[t]
\centering
\includegraphics[width=0.92\columnwidth]{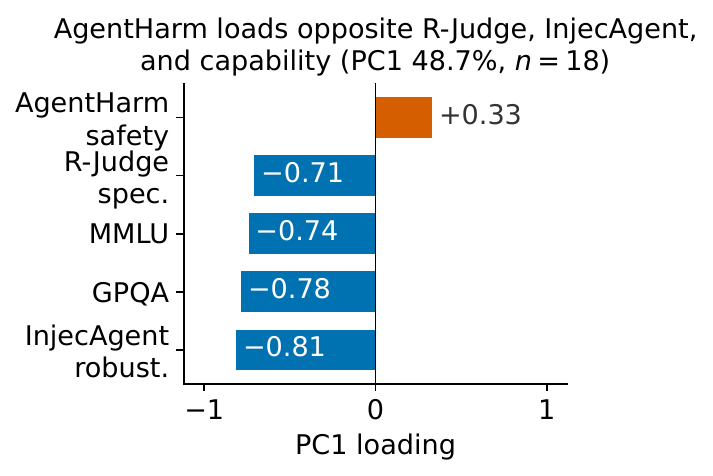}
\caption{\textbf{The first principal component groups R-Judge, InjecAgent, and capability, with
AgentHarm in the opposite direction.} PC1 loadings on the $n{=}18$ panel with harmonized anchors
($48.7\%$ of variance). AgentHarm has the opposite sign from the other four measures in $17/18$
leave-one-endpoint fits but $7/8$ leave-one-organization-out fits; deleting Meta changes its
oriented loading to $-0.10$. The sign of a principal component is arbitrary; only the relative
directions matter. We treat this structure as suggestive because the median between-benchmark
correlation is only $0.22$ (\S\ref{sec:rq1}).}
\label{fig:loadings}
\end{figure}

Figures~\ref{fig:reversal} and~\ref{fig:loadings} visualize the two RQ1 results: the dissolution
of the small-panel R-Judge--AgentHarm trade-off, and an exploratory first-component description of
how the scores covary.

An exploratory PCA gives a weaker description (Fig.~\ref{fig:loadings}): its first
component ($48.7\%$ variance) places R-Judge specificity, InjecAgent robustness, and both
capability anchors in one direction and AgentHarm safety in the other ($+0.33$). The direction
holds in $17/18$ model deletions but only $7/8$ organization deletions; dropping Meta changes the
oriented AgentHarm loading to $-0.10$, and PC1 ranges from $38.4\%$ to $52.7\%$. The PCA describes this roster and changes when Meta is omitted.

\paragraph{Planned PCA checks.} On harmonized anchors, MMLU loads
$0.74\geq0.6$ (versus $0.42$ with provider-reported anchors), and permuted matrices fall within the
parallel-analysis null; the convergent AgentDojo check is inconclusive at $n{=}5$. More importantly,
a rank-based PCA keeps the loading directions but drops PC1 to $40.8\%$ and Horn retains zero
factors (App.~\ref{app:rq1}); the median between-benchmark correlation is only $0.22$. We therefore
do not infer a latent safety factor. Table~\ref{tab:matrix} shows the ranking disagreement without using PCA.

\paragraph{Correlation-convention sensitivity.} The PCA above operates on the Pearson correlation
matrix of the standardized scores. A rank-based (Spearman) PCA keeps the loading structure (anchors
$-0.72$/$-0.78$, AgentHarm opposite at $+0.38$) and the recorded positive check still passes,
but PC1 falls from $48.7\%$ to $40.8\%$ of variance and Horn's parallel analysis retains no factor.
The recorded $40\%$ component-variance heuristic is met under both conventions, and the negative
check passes as executed: $200$ column-permuted
matrices give median first eigenvalue $1.71$ against the parallel-analysis $95$th-percentile
threshold of $2.18$ (exceedance $2\%$). Nevertheless, the zero-factor Horn result prevents the
heuristic from licensing a latent-factor claim. PCA is descriptive only; the primary RQ1 result is
the correlation-free ranking disagreement of Table~\ref{tab:matrix}.

\paragraph{Organization-deletion sensitivity.} On natural pairwise intersections, the full / LOO-
organization ranges are R-Judge--InjecAgent $+0.30/[+0.03,+0.48]$ ($n{=}21$, nine organizations),
R-Judge--AgentHarm $+0.02/[-0.22,+0.39]$ ($n{=}18$, eight), and
InjecAgent--AgentHarm $-0.33/[-0.51,-0.23]$ ($n{=}19$, eight). Across the eight PCA folds, PC1
explains $38.4$--$52.7\%$ and the oriented AgentHarm loading spans $-0.10$ to $+0.54$. These
folds support benchmark-dependent endpoint ordering, but not a population-stable factor geometry.

\subsection{R-Judge metric sensitivity (detail for \S\ref{sec:rq1}--\S\ref{sec:rq3})}
\label{app:metric}

Because R-Judge's headline F1 has no explicit true-negative term, while specificity (used for R-Judge in
the cross-instrument and external-variable analyses) ignores true positives, we
recompute every R-Judge-involving result under four metric choices on the paper's own panels
(Table~\ref{tab:metricsens}). The conclusions the paper leans on are metric-robust; the one
metric-sensitive cell is the R-Judge--misalignment partial, which the paper already treats
as exploratory.

\begin{table*}[t]
\centering
\small
\begin{tabular}{@{}lrrrrrrr@{}}
\toprule
 & \multicolumn{2}{c}{RQ1 ($n{=}18$)} & \multicolumn{2}{c}{RQ2 ($n{=}20$)} & \multicolumn{3}{c}{RQ3 partials} \\
\cmidrule(lr){2-3}\cmidrule(lr){4-5}\cmidrule(lr){6-8}
R-Judge metric & reversal & PC1 & MMLU & GPQA & $\tau^2$ & misalign.\ & jailbr.\ \\
\midrule
specificity & $+0.02$ & $-0.71$ & $+0.16$ & $+0.07$ & $-0.09$ & $+0.41$ & $-0.11$ \\
recall & $-0.05$ & $-0.14$ & $+0.40$ & $+0.32$ & $+0.15$ & $-0.46$ & $-0.05$ \\
balanced acc & $-0.09$ & $-0.86$ & $+0.71$ & $+0.49$ & $+0.19$ & $+0.19$ & $-0.20$ \\
F1 (official) & $-0.29$ & $-0.75$ & $+0.76$ & $+0.62$ & $+0.32$ & $-0.29$ & $-0.17$ \\
\bottomrule
\end{tabular}
\caption{R-Judge metric sensitivity under specificity, recall, balanced accuracy, and F1. RQ3 entries are partial Spearman correlations controlling the capability composite ($\tau^2$ $n{=}19$, misalignment $n{=}18$, jailbreak $n{=}20$). Under every metric, the RQ1 small-panel reversal does not return and no task-success or jailbreak partial is significant. The exploratory misalignment estimate is not stable: $+0.41$ under specificity and $-0.46$ under recall (both n.s.).}
\label{tab:metricsens}
\end{table*}

\subsection{Robustness of the capability crossover (detail for \S\ref{sec:rq3})}
\label{app:robustness}

\paragraph{Partial-correlation convention.} Partials residualize midranks on the harmonized
capability composite and Spearman-correlate the residual series. For the natural panels,
endpoint-unit $p$-values permute the outcome's capability-rank residuals ($99{,}999$ draws); this
constructs a capability-adjusted null rather than treating an ordinary case-bootstrap sign fraction
as a $p$-value. Under the classic alternative (Pearson on rank residuals) on the common $n{=}18$
panel, the two exploratory misalignment estimates are $+0.49$ (R-Judge) and $+0.41$
(InjecAgent), versus $+0.41/+0.47$ under our convention, and AgentHarm--jailbreak is $+0.74$ (vs.\ $+0.75$).
Table~\ref{tab:rq3pop} reports all natural-panel permutation values and exhaustive organization-cluster
intervals; no qualitative conclusion changes under either estimator.

\begin{figure}[t]
\centering
\includegraphics[width=0.92\columnwidth]{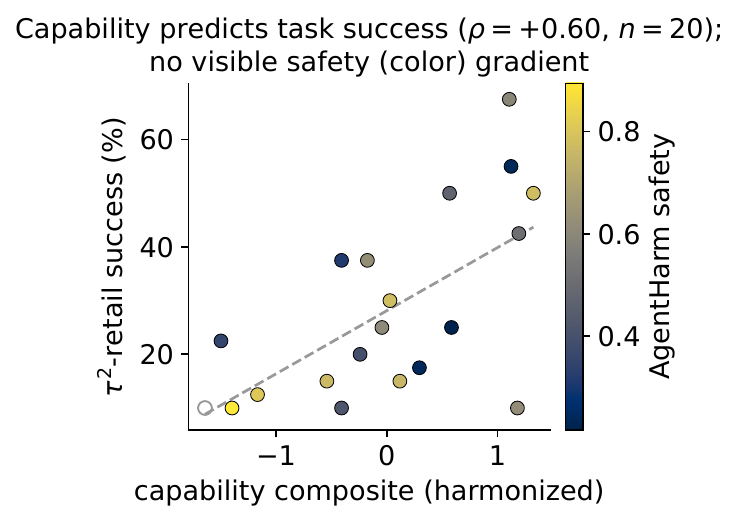}
\caption{\textbf{Against task success, the composite is associated and AgentHarm shows no detectable incremental association.}
$\tau^2$-retail success against the harmonized capability composite ($\rho=+0.60$,
$n{=}20$; gray trend). Color is AgentHarm safety (open circle: no AgentHarm score): safe and
unsafe models sit on both sides of the trend with no visible safety gradient after adjusting for the
composite, consistent with the nonsignificant partial in Table~\ref{tab:rq3}; the wide interval
(App.~\ref{app:rq3pop}) does not rule out an incremental association.}
\label{fig:rq3}
\end{figure}

Interaction $\Delta{=}{-}1.00$ (endpoint-bootstrap 95\% percentile interval $[-1.48,-0.49]$). Leave-one-out
$\Delta\in[-1.16,-0.90]$. Leave-one-organization-out, per omitted organization: Amazon $-0.96$, Anthropic
$-1.01$, Cohere $-0.90$, DeepSeek $-1.13$, Google $-0.98$, Meta $-0.96$, Mistral $-0.93$, OpenAI
$-1.16$, Qwen $-0.95$ (the range coincides with leave-one-out because Cohere contributes one model
and OpenAI, the largest group, gives the widest swing). Organization-clustered bootstrap:
$\Delta\in[-1.49,-0.59]$; all valid resamples are negative, which we report as conditional
sensitivity rather than a null-hypothesis test. Subsampling stability (draws of each size):
size-10 median $-0.95$, $100\%$ negative, $97\%\leq-0.5$, $0\%$ sign-flips; size-12
$-0.96$/$100\%$/$99\%$/$0\%$; size-14 $-0.97$/$100\%$/$100\%$/$0\%$ (contrast: the RQ1
R-Judge--AgentHarm correlation reaches $|\rho|{\geq}0.5$ in $\sim25\%$ of size-7 draws from
this fixed roster, a descriptive subsampling result).
Independent-grader checks:
the misalignment classifier reproduces at $\rho=0.97$ (Claude-3-Haiku vs.\ GPT-4o), the jailbreak judge
at $\rho=0.98$ (Gemini vs.\ gpt-4o), and a Gemini-judged refusal rate over the cached AgentHarm completions correlates $0.69$ with
AgentHarm safety ($n{=}19$). The refusal rate is a different quantity: the official scalar ignores the
refusal flag. Substituting it for AgentHarm safety gives an AgentHarm--jailbreak partial of $+0.67$
(vs.\ $+0.72$). Per-jailbreak AgentHarm$\to$jailbreak partials (within the three-template
suite): AIM $+0.44$, refusal-suppression $+0.70$, prefix-injection $+0.62$; the original
\emph{single}-AIM criterion gave $+0.50$.

\paragraph{Misalignment 3-scenario battery (exploratory).} Adding a third scenario (\emph{murder}; our
classifier, 22 models) to the blackmail$+$leaking criterion gives an
R-Judge misalignment-versus-jailbreak contrast of $\Delta=+0.53$ (endpoint-bootstrap percentile
interval $[0.00,+1.02]$, $n{=}20$); this differs from the misalignment-versus-task-success
$\Delta=+0.50$ in \S\ref{sec:rq3}. The point estimate is close to the two-scenario value; because this
post-hoc extension changes the uncertainty boundary without strengthening the estimated contrast,
we do not lean on it. Murder is also a severe harm that harm-compliance forecasts
(AgentHarm$\to$battery $+0.43$), diluting construct specificity, so we keep blackmail$+$leaking
primary.

\ifdefined\workshopversion
\subsection{The 2026 model-panel expansion (\texorpdfstring{n$\approx$41}{n approximately 41}) and organization-group differences (detail for
\S\ref{sec:rq3})}
\else
\subsection{The 2026 model-panel expansion (n$\approx$41) and organization-group differences (detail for
\S\ref{sec:rq3})}
\fi
\label{app:n40}
The expansion was a two-stage follow-up, each stage planned in a written note. A July 7
note fixed a gate that treated persistence or strengthening as a secondary expectation. After
observing that the gate weakened the estimate, a separate July 9 note fixed 17 further
candidates, telemetry-only eligibility rules, and a new falsifiable pooled-panel prediction before
collecting those candidates' outcomes. The July 9 stage was planned after the July 7 results, so the two stages were not planned together as one confirmatory study. We grew the panel to $41$
anchored models spanning $12$
model-developing organizations,
using the same chat/scenario harnesses as the original panel. The two criteria require no tool loop;
new models simply skip the tool-calling audit benchmarks. Exclusions were triggered only by the
specified telemetry rules, irrespective of outcome direction: Kimi-K2.6 and both Nemotron-3 models returned usable anchor
answers on fewer than $50\%$ of items or produced unbounded reasoning traces;
MiniMax-M3 fell below the misalignment engagement threshold because of truncated outputs, not low
harm, but was retained on the
single-turn jailbreak criterion, so the misalignment analysis uses $n{=}40$ whereas the jailbreak
analysis uses $n{=}41$.
Gemini-2.5-Pro yielded no valid anchor; GLM-4.5 yielded MMLU but not GPQA. Both encountered routing
incompatibilities, so only GLM-4.5 appears as a partial row in Fig.~\ref{fig:scale}. These exclusions reflect incompatibilities between the harness and some model routes.

\paragraph{Execution record.}
The authors' internal execution record places the adaptive expansion on July 7--12, 2026 through
OpenRouter's OpenAI-compatible endpoint; the public artifact does not independently timestamp this
chronology (App.~\ref{app:lineage}). It used Inspect AI 0.3.242, OpenAI 2.44.0, and
\texttt{inspect\_evals} commit
\texttt{a3decfa5}. MMLU uses temperature 0; GPQA temperature is not explicitly set and therefore
follows the provider default. Misalignment uses 25 rollouts in each of four scenario/goal cells,
temperature 1 for non-reasoning models and no explicit temperature for reasoning models, with a
fixed GPT-4o classifier rubric. Jailbreak uses the first 50 StrongREJECT prompts under each of three
published templates, temperature 0.75 for non-reasoning models and no explicit temperature for
reasoning models, with a fixed GPT-4o judge rubric. Both evaluator roles use the fixed request
identifier \texttt{openai/openai/gpt-4o}; OpenRouter routing and provider snapshots were not pinned.
Canonical provider slugs and telemetry gates are frozen in the run configuration.

\paragraph{Primary expansion result: the capability--misalignment correlation weakens.}
With the pooled capability scaling frozen before the stage split, the original 21 models give
$\rho=-0.452$, the six July 7 gate-only models $+0.541$, the 13 July 9-only models $+0.008$,
all 19 new models $+0.068$, and the pooled 40-model set $-0.163$. The July 9-only
estimate is compatible with attenuation but imprecise; the pooled estimate includes the
observations that motivated the follow-up and is not fresh independent confirmation.
At the paper's rounded reporting precision, capability$\to$misalignment-safety moves from
$-0.44$ ($n{=}21$, endpoint-bootstrap 95\% percentile interval $[-0.85,+0.09]$) to
$-0.16$ ($n{=}40$, interval $[-0.54,+0.22]$,
leave-one-out $[-0.25,-0.12]$); both intervals include zero. The second-stage note predicted
$\rho$ between $-0.20$ and $+0.25$ with an interval excluding $-0.44$, and said the prediction would
fail if $\rho\leq-0.35$ at $p<0.05$. The estimate falls in the predicted range and the failure
condition was not met, but the interval does not exclude $-0.44$; the note was not independently
timestamped before collection.
The estimate is weaker, but its interval still includes $-0.44$. Three direct
tests of the change itself (accounting for the original panel being nested in the expanded one):
a stratified nested bootstrap gives $\Delta\rho=+0.27$, $95\%$ CI $[-0.25,+0.78]$
(jailbreak $+0.26$, CI $[-0.14,+0.69]$); the 2026-only models alone give
$\rho=+0.07$ versus the original $-0.44$; and size-21 subsets
of the expanded pool matched to the original panel's organization histogram give median $\rho=-0.17$
(IQR $[-0.27,-0.08]$), i.e.\ organization composition alone does not reproduce the original value. So
organization mix alone does not explain the movement; generational change is compatible with the
result but is not established, and both direct change intervals include zero. The estimate depends on the model population. A ceiling does not explain all remaining
variation: among
models with similar measured capability, safety varies widely across organizations (composite ${\approx}0.8$:
o3-mini $0.99$, Grok-4.3 $0.70$, DeepSeek-V3 $0.19$). Because $\tau^2$ task success was not
re-collected, we update this correlation, not the interaction $\Delta$.

\paragraph{Organization-group analysis.} With $\geq4$ anchored
models each from six organizations we can compare capability and safety within organizations.
Within-organization capability--safety rank correlations vary in sign and magnitude, with $n$-weighted
means near zero;
for the seven groups with at least three endpoints, fixed-roster between-group shares are $0.508$
(misalignment) and $0.674$ (jailbreak), with df-adjusted Kruskal--Wallis
$\varepsilon^2=0.34$ and $0.57$ (Table~\ref{tab:n40}, Fig.~\ref{fig:developer}).
Leave-one-organization-out $\varepsilon^2$ ranges are $0.27$--$0.39$ and $0.41$--$0.69$,
respectively; dropping near-ceiling Anthropic gives $0.27$ and $0.54$. We treat these as
fixed-roster descriptions, not population tests: model endpoints are not exchangeable over
organization labels, and a synthetic calibration of a label-permutation statistic does not repair
that estimand. Capability is also matched only on a static-knowledge composite, within-organization
$n$ is $4$--$8$, and some organizations are range-restricted. Follow-up studies can test these differences within organization strata.

\begin{center}\small
\begin{tabular}{lccc}
\toprule
 & $n$ & misalign.\ $\rho$ & jailbreak $\rho$ \\
\midrule
\multicolumn{4}{l}{\emph{within-organization slopes:}} \\
Anthropic & 5 & $+0.71$ & $+0.67$ \\
OpenAI    & 8 & $+0.59$ & $+0.64$ \\
Qwen      & 5 & $-0.90$ & $+0.80$ \\
Meta      & 4 & $-0.40$ & $-0.40$ \\
Google    & 4 & $-0.60$ & $-0.40$ \\
Mistral   & 4 & $-1.00$ & $-0.40$ \\
\emph{$n$-weighted mean} & & $-0.14$ & $+0.26$ \\
\midrule
\multicolumn{4}{l}{\emph{organization-group description (group $n{\geq}3$; total $n{=}33$):}} \\
between-group share & & $0.508$ & $0.674$ \\
\ \ df-adjusted $\varepsilon^2$ & & $0.34$ & $0.57$ \\
\ \ leave-one-organization-out $\varepsilon^2$ & & $0.27$--$0.39$ & $0.41$--$0.69$ \\
\bottomrule
\end{tabular}
\end{center}
\captionof{table}{Organization-group differences on the $n{\approx}41$ expansion. Within an organization,
capability--safety rank correlations vary in sign and magnitude; their $n$-weighted means are $-0.14$
(misalignment) and $+0.26$ (jailbreak) (top). After rank-residualizing
safety on the static capability composite, the bottom rows describe between-group concentration on
the fixed roster; $\varepsilon^2$ is an effect-size summary, not a population test or causal share.}
\label{tab:n40}

\paragraph{Expanded-panel jailbreak result.} We rebuilt the jailbreak criterion on the
expanded panel (three published templates re-scored for every new model). Capability$\to$jailbreak-safety
rises from $+0.08$ ($n{=}21$) to $+0.34$ ($n{=}41$; endpoint-bootstrap 95\% percentile interval
$[+0.03,+0.60]$, leave-one-out $[+0.30,+0.39]$), while the direct panel-change interval
$[-0.14,+0.69]$ includes zero. Within-organization
slopes remain heterogeneous (weighted mean $+0.26$), while models with similar measured capability
vary across organizations (fixed-roster $\varepsilon^2{=}0.57$; LOO-organization range
$0.41$--$0.69$). Across all 41 models,
jailbreak safety spans $0.09$--$0.99$ (Mistral-Large $0.09$, DeepSeek-V3 $0.37$, Anthropic/OpenAI
$0.98$--$0.99$); among models with composite $>0.4$ it spans $0.45$--$0.99$. Neither the sign nor magnitude of capability$\to$safety is stable
across criteria or panels. An earlier protocolized R-Judge--misalignment continuation fell from
$+0.41$ (common panel, $n{=}18$) to $+0.23$ (all eligible models, $n{=}25$), below its $0.30$ threshold; of eight attempted gate
candidates, seven had complete records, six also had both anchors, and five additionally passed
the pre-recorded R-Judge-validity rule and entered the continuation (App.~\ref{app:lineage}).

\end{document}